\documentclass[10pt,twocolumn,letterpaper]{article}

\usepackage{cvpr}              

\usepackage{graphicx}
\usepackage{amsmath}
\usepackage{amssymb}
\usepackage{booktabs}

\usepackage[accsupp]{axessibility}

\usepackage{xcolor} 
\usepackage{times}
\usepackage{epsfig}
\usepackage{graphicx}
\usepackage{amsmath}
\usepackage{amssymb}
\usepackage[bb=dsserif]{mathalpha}
\usepackage{bm}

\usepackage{booktabs,colortbl,tabularx}
\usepackage{pifont}%

\usepackage{mathtools}
\usepackage{commath}
\usepackage{algorithm}
\usepackage[noend]{algpseudocode}

\usepackage{multirow}
\usepackage{comment} 
\usepackage{enumitem}

\definecolor{Gray}{gray}{0.9}

\newcommand{\cmark}{\ding{51}}%
\newcommand{\xmark}{\ding{55}}%

\definecolor{battleshipgrey}{rgb}{0.52, 0.52, 0.51}

\usepackage[pagebackref,breaklinks,colorlinks]{hyperref}

\usepackage[capitalize]{cleveref}
\crefname{section}{Sec.}{Secs.}
\Crefname{section}{Section}{Sections}
\Crefname{table}{Table}{Tables}
\crefname{table}{Tab.}{Tabs.}

\def\confName{CVPR}
\def\confYear{2023}

\begin{document}

\title{Audio-Visual Grouping Network for Sound Localization from Mixtures}

\author{%
  Shentong Mo\\
  Carnegie Mellon University
  \and
  Yapeng Tian\thanks{Corresponding author.} \\
  University of Texas at Dallas 
}

\maketitle

\begin{abstract}

Sound source localization is a typical and challenging task that predicts the location of sound sources in a video.
Previous single-source methods mainly used the audio-visual association as clues to localize sounding objects in each image.
Due to the mixed property of multiple sound sources in the original space, there exist rare multi-source approaches to localizing multiple sources simultaneously, except for one recent work using a contrastive random walk in the graph with images and separated sound as nodes.
Despite their promising performance, they can only handle a fixed number of sources, and they cannot learn compact class-aware representations for individual sources. 
To alleviate this shortcoming, in this paper, we propose a novel audio-visual grouping network, namely AVGN, that can directly learn category-wise semantic features for each source from the input audio mixture and image to localize multiple sources simultaneously.
Specifically, our AVGN leverages learnable audio-visual class tokens to aggregate class-aware source features.
Then, the aggregated semantic features for each source can be used as guidance to localize the corresponding visual regions.
Compared to existing multi-source methods, our new framework can localize a flexible number of sources and disentangle category-aware audio-visual representations for individual sound sources.
We conduct extensive experiments on MUSIC, VGGSound-Instruments, and VGG-Sound Sources benchmarks.
The results demonstrate that the proposed AVGN can achieve state-of-the-art sounding object localization performance on both single-source and multi-source scenarios.
Code is available at \url{https://github.com/stoneMo/AVGN}.

\end{abstract}


\vspace{-1em}
\section{Introduction}

When we hear a dog barking, we are naturally aware of where the dog is in the room due to the strong correspondence between audio signals and visual objects in the world.
In the meanwhile, we are capable of separating individual sources from a mixture of multiple sources in the daily environment.
This human perception intelligence attracts many researchers to explore audio-visual joint learning for visual sound source localization.

\begin{figure}[t]
\centering
\includegraphics[width=0.99\linewidth]{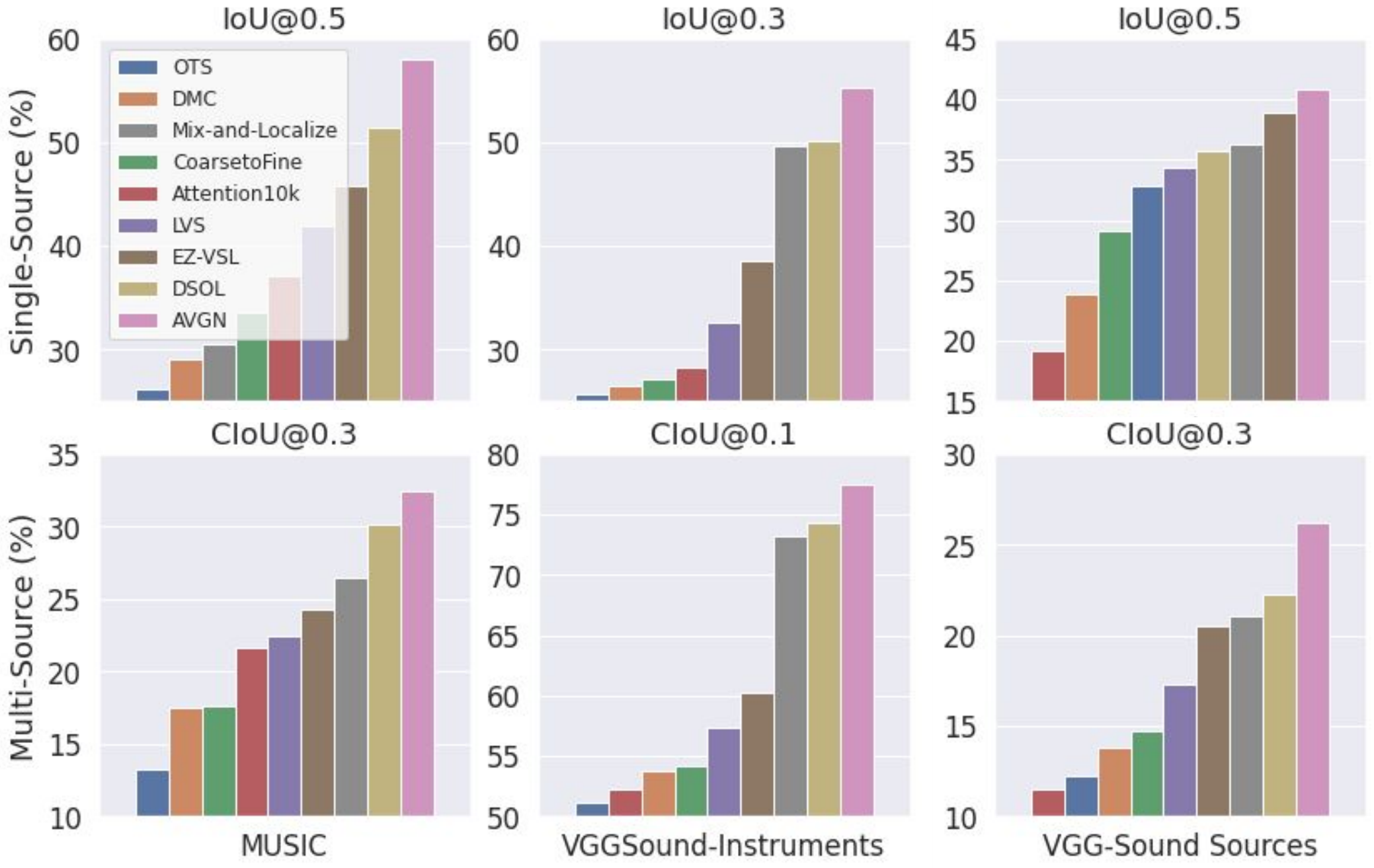}
\vspace{-0.5em}
\caption{Comparison of our AVGN with state-of-the-art methods on single-source (Top Row) and multi-source (Bottom Row) sound localization on MUSIC~\cite{zhao2018the}, VGGSound-Instruments~\cite{hu2022mix}, and VGG-Sound Sources~\cite{chen2021localizing} benchmarks. }
\label{fig: title_img}
\vspace{-2.0em}
\end{figure}

Visual sound source localization is a typical and challenging task that predicts the location of sound sources in a video.
To tackle this problem, early single-source methods~\cite{hu2019deep,Senocak2018learning,Afouras2020selfsupervised,chen2021localizing,arda2022learning,mo2022EZVSL,mo2022SLAVC} mainly used the audio-visual association as clues to localize sounding objects in the frame.
Typically, Attention10k~\cite{Senocak2018learning} introduced a two-stream architecture for audio and images to localize a sound source in the image using an attention mechanism.
Based on the attention, Afouras \textit{et al.}~\cite{Afouras2020selfsupervised} incorporated the optical flow for more accurate localization in a video.
To explicitly learn discriminative audio-visual corresponding fragments, LVS~\cite{chen2021localizing} proposed hard sample mining with a differentiable threshold-based contrastive loss, while HardPos~\cite{arda2022learning} utilized hard positives from negative pairs in the loss.
More recently, EZVSL~\cite{mo2022EZVSL} developed a multiple-instance contrastive learning framework on the most aligned regions corresponding to the audio.
SLAVC~\cite{mo2022SLAVC} adopted momentum encoders and extreme visual dropout to address overfitting and silence issues in single-source sound localization.
However, those baselines are based on the single-source sound as input and they achieve worse performance for sound localization from mixtures.
In this work, we will solve the problem in our approach by extracting disentangled and compact representations with learnable audio-visual class tokens as guidance for sound localization.

Since multiple sound sources are mixed in the original space, recent researchers have tried to explore diverse pipelines to localize multiple sources on frames from a sound mixture.
This multi-source task requires the model to associate individual sources separated from the mixture with each frame.
Qian \textit{et al.}~\cite{qian2020multiple} leveraged a two-stage framework to capture cross-modal feature alignment between sound and vision representations in a coarse-to-fine manner.
DSOL~\cite{hu2020dsol} introduced a two-stage training framework to tackle with silence in category-aware sound source localization.
More recently, Mix-and-Localize~\cite{hu2022mix} proposed to use a contrastive random walk in the graph with images and separated sounds as nodes, where a random walker was trained to walk from each audio node to an image node with audio-visual similarity as the transition probability.
Despite their promising performance, they can only handle a fixed number of sources and they cannot learn compact class-aware representations for individual sources. 
In contrast, we can support a flexible number of sources as input and learn class-aware representations for each source.

The main challenge is that sounds are naturally mixed in the audio space.
This inspires us to disentangle the individual semantics for each source from the mixture to guide source localization.
To address the problem, our key idea is to disentangle individual source representation using audio-visual grouping for source localization, which is different from existing single-source and multi-source methods. 
During training, we aim to learn audio-visual category tokens to aggregate category-aware source features from the sound mixture and the image, where separated high-level semantics for individual sources are learned.

To this end, we propose a novel audio-visual grouping network, namely AVGN, that can directly learn category-wise semantic features for each source from the input audio mixture and frame to localize multiple sources simultaneously.
Specifically, our AVGN leverages learnable audio-visual class tokens to aggregate class-aware source features.
Then, the aggregated semantic features for each source will serve as guidance to localize the corresponding visual regions.
Compared to previous multi-source baselines, our new framework can support a flexible number of sources and shows the effectiveness of learning compact audio-visual representations with category-aware semantics.

Empirical experiments on MUSIC and VGGSound-Instruments benchmarks comprehensively demonstrate the state-of-the-art performance against previous single-source and multi-source baselines.
In addition, qualitative visualizations of localization results vividly showcase the effectiveness of our AVGN in localizing individual sources from mixtures.
Extensive ablation studies also validate the importance of category-aware grouping and learnable audio-visual class tokens in learning compact representations for sound source localization.

Our main contributions can be summarized as follows:
\begin{itemize}
    \item We present a novel Audio-Visual Grouping Network, namely AVGN, to disentangle the individual semantics from sound mixtures and images to guide source localization.
    \item We introduce learnable audio-visual class tokens and category-aware grouping in sound localization to aggregate category-wise source features with explicit high-level semantics.
    \item Extensive experiments comprehensively demonstrate the state-of-the-art superiority of our AVGN over previous baselines on both single-source and multi-source sounding object localization.

\end{itemize}

\vspace{-0.5em}
\section{Related Work}

\begin{figure*}[t]
    \centering
    \includegraphics[width=0.9\linewidth]{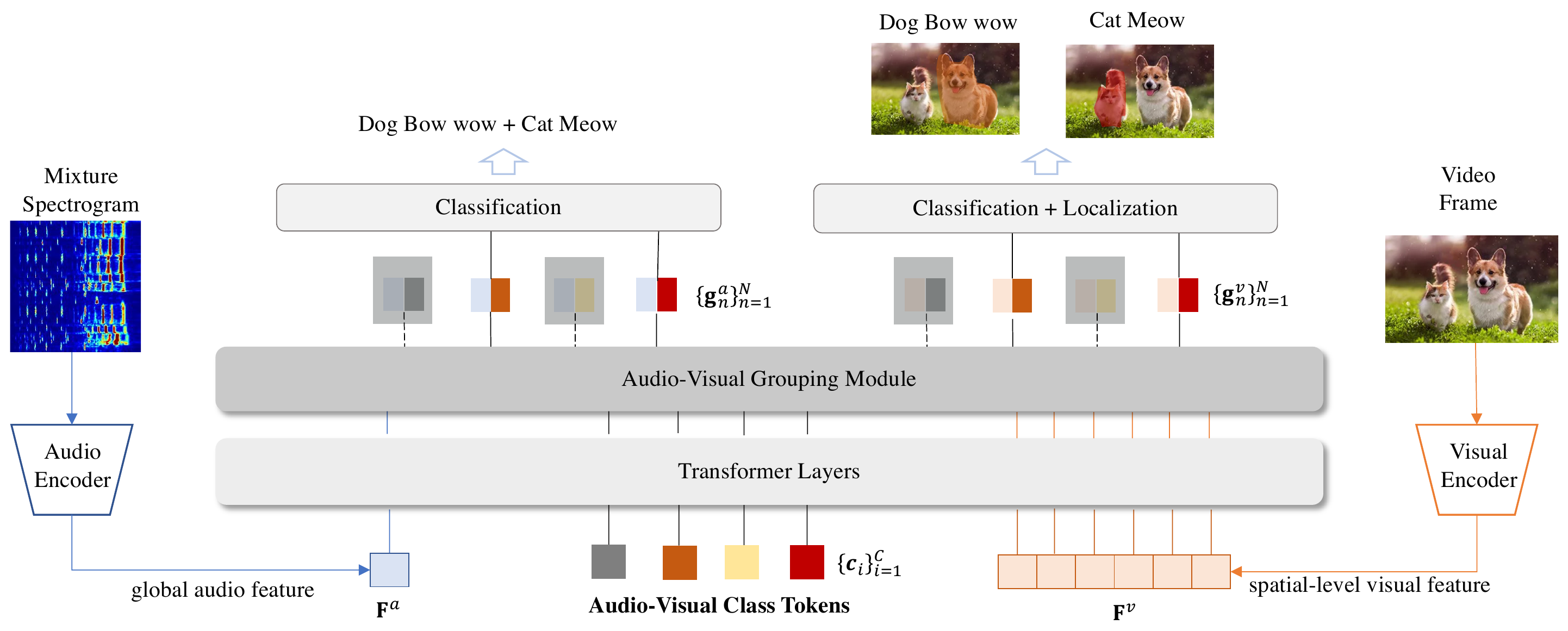}
    \vspace{-0.5em}
    \caption{Illustration of the proposed Audio-Visual Grouping Network (AVGN).
    The Audio-Visual Grouping module takes as global audio $\mathbf{F}^a=\mathbf{f}^a$ of the mixture spectrogram, spatial-level visual features $\mathbf{F}^v=\{\mathbf{f}^v_p\}_{p=1}^P$ of the video frame from each encoder and learnable audio-visual class tokens $\{\mathbf{c}_i\}^C_{i=1}$ of for $C$ categories in the semantic space to generate disentangled class-aware audio-visual representations $\{\mathbf{g}_n^a\}^N_{n=1}, \{\mathbf{g}_n^v\}^N_{n=1}$ for $N$ sources.
    Note that $N$ source embeddings are chosen from $C$ categories according to the ground-truth class.
    Finally, two classification layers composed of an FC layer and a sigmoid function are separately used to predict audio and video categories, and localization maps are generated by the cosine similarity between audio-visual class-aware embeddings.
    }
    \label{fig: main_img}
    \vspace{-1em}
\end{figure*}

\noindent\textbf{Audio-Visual Joint Learning.}
Audio-visual joint learning has been addressed in many previous works~\cite{aytar2016soundnet,owens2016ambient,Arandjelovic2017look,korbar2018cooperative,Senocak2018learning,zhao2018the,zhao2019the,Gan2020music,Morgado2020learning,Morgado2021robust,Morgado2021audio,hershey2001audio,ephrat2018looking,hu2019deep} to learn the audio-visual correlation between two distinct modalities from videos.
Such cross-modal alignments are beneficial for many audio-visual tasks, such as audio-event localization~\cite{tian2018ave,lin2019dual,wu2019dual,lin2020audiovisual}, audio-visual spatialization~\cite{Morgado2018selfsupervised,gao20192.5D,Chen2020SoundSpacesAN,Morgado2020learning}, audio-visual navigation~\cite{Chen2020SoundSpacesAN,chen2021waypoints,chen22soundspaces2} and audio-visual parsing~\cite{tian2020avvp,wu2021explore,lin2021exploring,mo2022multimodal}.
In this work, our main focus is to learn compact audio-visual representations for localizing individual sources on images from sound mixtures, which is more demanding than the tasks aforementioned above.

\noindent\textbf{Audio-Visual Source Separation.}
Audio-visual source separation aims to separate individual sound sources from the audio mixture given the image with sources on it.
In recent years, researchers~\cite{hershey2001audio,zhao2018the,ephrat2018looking,Gao2018learning,xu2019mpnet,tian2021cyclic,tzinis2020into} have tried to explore diverse pipelines to learn discriminative visual representations from images for source separation.
Zhao \textit{et al.}~\cite{zhao2018the} first proposed a ``Mix-and-Separate'' network to capture the alignment between pixels and the spectral components of audio for the reconstruction of each input source spectrogram.
With the benefit of visual cues, MP-Net~\cite{xu2019mpnet} utilized a recursive MinusPlus Net to separate all salient sounds from the mixture.
Tian \textit{et al.}~\cite{tian2021cyclic} used a cyclic co-learning framework with sounding object visual grounding to separate visual sound sources.
More recently, more types of modalities have been explored to boost the performance of audio-visual source separation, such as motion in SoM~\cite{zhao2019the}, gesture composed of pose and keypoints in MG~\cite{Gan2020mg}, and spatio-temporal visual scene graphs in AVSGS~\cite{Chatterjee2021visual}.
Different from them, we do not need to recover the audio spectrogram of individual sources from the mixture.
Instead, we leverage the category-aware representations of individual sources to localize the corresponding regions for each source, where learnable audio-visual class tokens are applied as the desirable guidance.

\noindent\textbf{Visual Sound Source Localization.}
Visual sound source localization is a typical and challenging problem that predicts the location of individual sound sources in a video.
Early works~\cite{hershey1999audio,fisher2000learning,kidron2005pixels} applied traditional machine learning approaches, such as statistical models~\cite{fisher2000learning} and canonical correlation analysis~\cite{kidron2005pixels} to learn low-level alignment between audio and visual representations.
With the success of deep neural nets, recent researchers~\cite{Senocak2018learning,hu2019deep,Afouras2020selfsupervised,qian2020multiple,chen2021localizing,arda2022learning,mo2022EZVSL,mo2022SLAVC} explored many architectures to learn the audio-visual correspondence for localizing single-source sounds.
Attention10k~\cite{Senocak2018learning} localized a sound source in the image using a two-stream architecture with an attention mechanism.
Hard sample mining was introduced in LVS~\cite{chen2021localizing} to optimize a differentiable threshold-based contrastive loss for predicting discriminative audio-visual correspondence maps.
More recently, a multiple-instance contrastive learning framework was proposed in EZVSL~\cite{mo2022EZVSL} to align regions with the most corresponding audio without negative regions involved.
\vspace{-0.1em}

Due to the natural mixed property of sounds in our environment, recent works~\cite{qian2020multiple,hu2020dsol,hu2022mix} also have explored different frameworks to localize multiple sources on frames from a sound mixture simultaneously.
DSOL~\cite{hu2020dsol} utilized a two-stage training framework to deal with silence for category-aware sound source localization.
More recently, a contrastive random walk model was trained in Mix-and-Localize~\cite{hu2022mix} to link each audio node with an image node using a transition probability of audio-visual similarity. 
While those single-source and multi-source approaches achieve promising performance in sound localization, they can only handle a fixed number of sources and they cannot learn discriminative class-aware representations for individual sources.
In contrast, we develop a fully novel framework to aggregate compact category-wise audio and visual source representations with explicit learnable source class tokens. 
To the best of our knowledge, we are the first to leverage an explicit grouping mechanism for sound source localization.
Our experiments in Section~\ref{sec:exp} also demonstrate the effectiveness of AVGN in both single-source and multi-source localization.

\vspace{-0.5em}

\section{Method}

\vspace{-0.5em}

Given an image and a mixture of audio, our target is to localize individual sound sources on the image. 
We propose a novel Audio-Visual Grouping Network named AVGN for disentangling individual semantics from the mixture and image, which mainly consists of two modules, Audio-Visual Class Tokens in Section~\ref{sec:avct} and Audio-Visual Grouping in Section~\ref{sec:avg}.

\vspace{-0.5em}
\subsection{Preliminaries}
\vspace{-0.5em}
In this section, we first describe the problem setup and notations, and then revisit the multiple-instance contrastive learning in EZVSL~\cite{mo2022EZVSL} for single-source localization.

\noindent\textbf{Problem Setup and Notations.}
Given a mixture spectrogram and an image, our goal is to localize $N$ individual sound sources in the image spatially.
For a video with $C$ source event categories, we have an audio-visual label, which is denoted as $\{y_i\}_{i=1}^C$ with $y_i$ for the ground-truth category entry $i$ as 1. 
During the training, we do not have bounding boxes and mask-level annotations. 
Therefore, we can only use the video-level label for the mixture spectrogram and image to perform weakly-supervised learning.

\noindent\textbf{Revisit Single-source Localization.}
To address the single-source localization problem, EZ-VSL~\cite{mo2022EZVSL} introduced a multiple-instance contrastive learning framework to align the audio and visual features at locations corresponding to sound sources.
Given a set of global audio feature $\mathbf{F}^a=\mathbf{f}^a\in\mathbb{R}^{1\times D}$ and spatial-level visual features spanning all locations in an image $\mathbf{F}^v = \{\mathbf{f}^v_p\}_{p=1}^P, \mathbf{f}^v_p\in\mathbb{R}^{1\times D}$, EZ-VSL applied the multiple-instance contrastive objective to align at least one location in the corresponding bag of visual features with the audio representation in the same mini-batch, which is defined as:
\begin{equation}\label{eq:micl}
    \mathcal{L}_{\mbox{baseline}} = 
    - \frac{1}{B}\sum_{b=1}^B \log \frac{
    \exp \left( \frac{1}{\tau} \mathtt{sim}(\mathbf{F}^a_b, \mathbf{F}^v_b) \right)
    }{
    \sum_{m=1}^B \exp \left(  \frac{1}{\tau} \mathtt{sim}(\mathbf{F}^a_b, \mathbf{F}^v_m)\right)}
\end{equation}
where the similarity $\mathtt{sim}(\mathbf{F}^a, \mathbf{F}^v)$ denotes the max-pooled audio-visual cosine similarity of $\mathbf{F}^a$ and $\mathbf{F}^v =\{\mathbf{f}^v_p\}_{p=1}^P$ across all $P$ spatial locations.
$B$ is the batch size, $D$ is the dimension size, and $\tau$ is a temperature hyper-parameter.

However, such a training objective will pose the main challenge for multi-source localization.
The global audio representation extracted from the mixture is mixed and thus they can not associate individual sources separated from the mixture with the corresponding regions.
To address the challenge, we are inspired by~\cite{xu2022groupvit} and propose a novel Audio-Visual Grouping Network that can learn to disentangle the individual semantics from the mixture and image to guide multi-source localization, as illustrated in Figure~\ref{fig: main_img}.

\vspace{-0.5em}
\subsection{Audio-Visual Class Tokens}\label{sec:avct}
\vspace{-0.5em}
In order to explicitly disentangle individual semantics from the mixed sound space and image, we introduce a novel learnable audio-visual class tokens $\{\mathbf{c}_i\}^C_{i=1}$ to help group semantic-aware information from audio-visual representations $\mathbf{f}^a, \{\mathbf{f}^v_p\}_{p=1}^P$, where $\mathbf{c}_i\in\mathbb{R}^{1\times D}$, $C$ is the total number of source classes, $P$ denotes the number of total locations in the spatial map.

With the categorical audio-visual tokens and raw representations, we first apply self-attention transformers $\phi^a(\cdot), \phi^v(\cdot)$ to aggregate global audio and spatial visual features from the raw input and align the features with the categorical token embeddings as:
\begin{equation}
\begin{aligned}
    & \hat{\mathbf{f}}^a, \{\hat{\mathbf{c}}_i^a\}_{i=1}^C = \{\phi^a(\mathbf{x}^a_{j}, \mathbf{X}^a, \mathbf{X}^a)\}_{j=1}^{1+C}, \\
    & \mathbf{X}^a = \{\mathbf{x}^a_{j}\}_{j=1}^{1+C} = [\mathbf{f}^a; \{\mathbf{c}_i\}_{i=1}^C]
\end{aligned}
\end{equation}
\begin{equation}
\begin{aligned}
    & \{\hat{\mathbf{f}}_p^v\}_{p=1}^P, \{\hat{\mathbf{c}}_i^v\}_{i=1}^C = \{\phi^v(\mathbf{x}^v_j, \mathbf{X}^v, \mathbf{X}^v)\}_{j=1}^{P+C}, \\
    & \mathbf{X}^v = \{\mathbf{x}^v_{j}\}_{j=1}^{P+C} = [\{\mathbf{f}_p^v\}_{p=1}^P; \{\mathbf{c}_i\}_{i=1}^C]
\end{aligned}
\end{equation}
where $[\ ;\ ]$ denotes the concatenation operator. 
$\hat{\mathbf{f}}^a, \hat{\mathbf{f}}_p^v, \hat{\mathbf{c}}_i^a, \hat{\mathbf{c}}_i^v\in\mathbb{R}^{1\times D}$, and and $D$ is the dimension of embeddings.
The self-attention operators $\phi^a(\cdot)$ is formulated as:
\begin{equation}
    \phi^a(\mathbf{x}_j^a, \mathbf{X}^a, \mathbf{X}^a) = \mbox{Softmax}(\dfrac{\mathbf{x}_j^a{\mathbf{X}^a}^\top}{\sqrt{D}})\mathbf{X}^a
\end{equation}
\begin{equation}
    \phi^v(\mathbf{x}_j^v, \mathbf{X}^v, \mathbf{X}^v) = \mbox{Softmax}(\dfrac{\mathbf{x}_j^v{\mathbf{X}^v}^\top}{\sqrt{D}})\mathbf{X}^v
\end{equation}
Then, in order to constrain the independence of each class token $\mathbf{c}_i$ in the audio-visual semantic space, we apply a fully-connected (FC) layer and add softmax operator to predict the individual source class probability: 
$\mathbf{e}_i = \mbox{Softmax}(\textsc{FC}(\mathbf{c}_i))$. 
Each audio-visual category probability is optimized by a cross-entropy loss
$\sum_{i=1}^C\mbox{CE}(\mathbf{h}_i, \mathbf{e}_i)$, where $\mbox{CE}(\cdot)$ is cross-entropy loss; $\mathbf{h}_i$ denotes a one-hot encoding with its element for the target category entry $i$ as 1.
Optimizing the loss will push the learned token embeddings to be category-aware and discriminative.

\begin{table*}[t]
	\renewcommand\tabcolsep{6.0pt}
	\centering
	\scalebox{0.8}{
		\begin{tabular}{l|ccc|ccc|cccc}
			\toprule
			\multirow{2}{*}{Method} & \multicolumn{3}{c|}{MUSIC-Solo} & \multicolumn{3}{c|}{VGGSound-Instruments} & \multicolumn{3}{c}{VGGSound-Single}  \\
			& AP(\%) & IoU@0.5(\%) & AUC(\%) &  AP(\%) & IoU@0.3(\%) & AUC(\%) &  AP(\%) & IoU@0.5(\%) & AUC(\%) \\ 	
			\midrule
			Attention10k~\cite{Senocak2018learning} & --   & 37.2 & 38.7 & -- & 28.3 & 26.1 & -- & 19.2 & 30.6 \\
               OTS~\cite{Arandjelovic2018ots} & 69.3 & 26.1 & 35.8 & 47.5 & 25.7 & 24.6 & 29.8 & 32.8 & 35.7 \\

		   DMC~\cite{hu2019deep} & --   & 29.1 & 38.0 & -- & 26.5 & 25.7 & -- & 23.9 & 27.6 \\
            CoarsetoFine~\cite{qian2020multiple} & 70.7 & 33.6 & 39.8  & 40.2 & 27.2 & 26.5 & 28.2 & 29.1 & 34.8 \\

			LVS~\cite{chen2021localizing} & 70.6 & 41.9 & 40.3 & 42.3 & 32.6 & 28.3 & 29.6 & 34.4 & 38.2 \\

			EZ-VSL~\cite{mo2022EZVSL}  & 71.5 & 45.8 & 41.2 & 43.8 & 38.5 & 30.6 & 31.3 & 38.9 & 39.5 \\

                Mix-and-Localize~\cite{hu2022mix}  & 68.6 & 30.5 & 40.8 & 44.9 & 49.7 & 32.3 & 32.5 & 36.3 & 38.9 \\

                DSOL~\cite{hu2020dsol} & --   & 51.4 & 43.7 & -- & 50.2 & 32.9 & -- & 35.7 & 37.2 \\

                AVGN (ours) & \textbf{77.2} & \textbf{58.1} & \textbf{48.5} & \textbf{50.5} & \textbf{55.3} & \textbf{36.7} & \textbf{35.3} & \textbf{40.8} & \textbf{42.3} \\

			\bottomrule
			\end{tabular}}
   \vspace{-0.5em}
   \caption{Quantitative results of single-source localization on MUSIC-Solo, VGGSound-Instruments, and VGGSound-Single datasets.}
   \label{tab: exp_sota_single}
   \vspace{-1.5em}
\end{table*}

\vspace{-0.5em}
\subsection{Audio-Visual Grouping}\label{sec:avg}
\vspace{-0.5em}
With the benefit of the aforementioned category-constraint objective, we propose a novel and explicit audio-visual grouping module composed of grouping blocks $g^a(\cdot), g^v(\cdot)$ to take the learned audio-visual source class tokens and aggregated features as inputs to generate category-aware audio-visual embeddings as:
\begin{equation}
\begin{aligned}
    \{\mathbf{g}_i^a\}_{i=1}^C  &= g^a(\{\hat{\mathbf{f}}^a, \{\hat{\mathbf{c}}_i^a\}_{i=1}^C),\\
    \{\mathbf{g}_i^v\}_{i=1}^C  &= g^v(\{\hat{\mathbf{f}}_p^v\}_{p=1}^P, \{\hat{\mathbf{c}}_i^v\}_{i=1}^C)
\end{aligned}
\end{equation}
During the phase of grouping, we merge all the audio-visual features from the same class token into a new class-aware audio-visual feature, by calculating the global audio similarity vector $\mathbf{A}^a\in\mathbb{R}^{1\times C}$ and spatial visual similarity matrix $\mathbf{A}^v\in\mathbb{R}^{P\times C}$ between audio-visual features and audio-visual class tokens via a softmax operation, which is formulated as
\begin{equation}
\begin{aligned}
    & \mathbf{A}^a_{i} = \mbox{Softmax}(W_q^a\hat{\mathbf{f}}^a \cdot W_k^a\hat{\mathbf{c}}_i^a), \\
    & \mathbf{A}^v_{p,i} = \mbox{Softmax}(W_q^v\hat{\mathbf{f}}_p^v \cdot W_k^v\hat{\mathbf{c}}_i^v)
\end{aligned}
\end{equation}
where $W_q^a, W_k^a\in\mathbb{R}^{D\times D}$ and $W_q^v, W_k^v\in\mathbb{R}^{D\times D}$ denote the learnable weights of linear projections for the features and class tokens of audio and visual modalities, separately.
With this global audio similarity vector and spatial visual similarity matrix, we compute the weighted sum of all global audio and spatial visual features assigned to generate the category-aware representations as:
\begin{equation}\label{eq:uni_group}
\begin{aligned}
    \mathbf{g}_i^a & = g^a(\hat{\mathbf{f}}^a, \hat{\mathbf{c}}_i^a)=\hat{\mathbf{c}}_i^a + W_o^a\dfrac{\mathbf{A}^a_{i}W_v^a\hat{\mathbf{f}}^a}{\mathbf{A}^a_{i}}, \\
    \mathbf{g}_i^v & = g^v(\{\hat{\mathbf{f}}_p^v\}_{p=1}^P, \hat{\mathbf{c}}_i^v) = \hat{\mathbf{c}}_i^v + 
    W_o^v\dfrac{\sum_{p=1}^{P}\mathbf{A}^v_{p,i}W_v^v\hat{\mathbf{f}}_p^v}{\sum_{p=1}^{P}\mathbf{A}^v_{p,i}}
\end{aligned}
\end{equation}
where $W_o^a, W_v^a\in\mathbb{R}^{D\times D}$ and $W_o^v, W_v^v\in\mathbb{R}^{D\times D}$ denote the learned weights of linear projections for output and value in terms of audio and visual modalities, separately. 
With class-aware audio-visual representations $\{\mathbf{g}_i^a\}_{i=1}^C, \{\mathbf{g}_i^v\}_{i=1}^C$ as the inputs, we use an FC layer and sigmoid operator on each modality to predict the binary probability: $p_i^a = \mbox{Sigmoid}(\textsc{FC}(\mathbf{g}_i^a)), p_i^v = \mbox{Sigmoid}(\textsc{FC}(\mathbf{g}_i^v))$ for $i$th class. 
By applying audio-visual source classes $\{y_i\}_{i=1}^C$ as the weak supervision and combining the class-constraint loss, we formulate an audio-visual grouping loss as:
\begin{equation}
    \mathcal{L}_{\mbox{group}} = \sum_{i=1}^C\{\mbox{CE}(\mathbf{h}_i, \mathbf{e}_i) + \mbox{BCE}(y_i, p_i^a) + \mbox{BCE}(y_i, p_i^v)\}.
\end{equation}
Since multiple audio sources could be in one mixture, we use binary cross-entropy loss: $\mbox{BCE}(\cdot)$ for each category to handle this multi-label classification problem.

With the help of the proposed class-constrained loss, we generate category-aware audio and visual representations $\{\mathbf{g}_i^a\}_{i=1}^C, \{\mathbf{g}_i^v\}_{i=1}^C$ for audio-visual alignment.
Note that global audio and visual representations for $N$ source embeddings $\{\mathbf{g}_n^a\}_{n=1}^N, \{\mathbf{g}_n^v\}_{n=1}^N$ are chosen from $C$ categories according to the corresponding ground-truth class. 
Therefore, the audio-visual similarity is calculated by max-pooling audio-visual cosine similarities of class-aware audio feature $\mathbf{g}_n^a$ and the spatial-level $\{\mathbf{f}^v_p\odot\mathbf{g}_n^v\}_{p=1}^P$ across all $P$ locations for $n$th source.
With this category-aware similarity, we formulate the new localization loss as:
\begin{equation}\label{eq:micl}
    \mathcal{L}_{\mbox{loc}} = 
    - \frac{1}{BN}\sum_{b=1}^B\sum_{n=1}^N \log \frac{
    \exp \left( \frac{1}{\tau} \mathtt{sim}(\mathbf{F}^a_{b,n}, \mathbf{F}^v_{b,n}) \right)
    }{
    \sum_{m=1}^B \exp \left(  \frac{1}{\tau} \mathtt{sim}(\mathbf{F}^c_{b,n}, \mathbf{F}^v_{m,n})\right)}
\end{equation}
where $\mathbf{F}^a_{b,n} = \mathbf{g}_n^a, \mathbf{F}^v_{b,n}=\{\mathbf{f}^v_p\odot\mathbf{g}_n^v\}_{p=1}^P$ denote the class-aware audio and visual features of $n$th source for $b$th sample in the mini-batch.
The overall objective of our model is simply optimized in an end-to-end manner as:
\begin{equation}
    \mathcal{L} = \mathcal{L}_{\mbox{loc}} + \mathcal{L}_{\mbox{group}}
\end{equation}
During inference, we follow the prior work~\cite{mo2022EZVSL} and use audio-visual cosine similarity map between class-aware audio-visual representations $\mathbf{g}_n^a, \{\mathbf{f}^v_p\odot\mathbf{g}_n^v\}_{p=1}^P$ to generate $n$th source localization map with $P$ locations.
Similarly to previous works~\cite{chen2021localizing,mo2022EZVSL,mo2022SLAVC}, the final localization map is generated through bilinear interpolation of the similarity map.

\begin{table*}[t]
	\renewcommand\tabcolsep{4.0pt}
	\centering
	\scalebox{0.7}{
		\begin{tabular}{l|cccc|cccc|ccccc}
			\toprule
			\multirow{2}{*}{Method} & \multicolumn{4}{c|}{MUSIC-Duet} & \multicolumn{4}{c|}{VGGSound-Instruments} &  \multicolumn{4}{c}{VGGSound-Duet}  \\
			& CAP(\%) & PIAP(\%) & CIoU@0.3(\%) & AUC(\%) & CAP(\%) & PIAP(\%) & CIoU@0.1(\%) & AUC(\%) & CAP(\%) & PIAP(\%) & CIoU@0.3(\%) & AUC(\%) \\ 	
			\midrule
			Attention10k~\cite{Senocak2018learning} & -- & -- & 21.6 & 19.6 & -- & -- & 52.3 & 11.7 & -- & -- & 11.5 & 15.2 \\

            OTS~\cite{Arandjelovic2018ots} & 11.6 & 17.7 & 13.3 & 18.5 & 23.3 & 37.8 & 51.2 & 11.2 & 10.5 & 12.7 & 12.2 & 15.8 \\

			DMC~\cite{hu2019deep} & --& --& 17.5 & 21.1 & -- & -- & 53.7 & 12.5 & -- & -- & 13.8 & 17.1 \\

   CoarsetoFine~\cite{qian2020multiple} & --& --& 17.6 & 20.6 & -- & -- & 54.2 & 12.9 & -- & -- & 14.7 & 18.5 \\

			LVS~\cite{chen2021localizing} & --& --& 22.5 & 20.9 & -- & -- & 57.3 & 13.3 & -- & -- & 17.3 & 19.5 \\

			EZ-VSL~\cite{mo2022EZVSL}  & --& --& 24.3 & 21.3 & -- & -- & 60.2 & 14.2 & -- & -- & 20.5 & 20.2 \\

                Mix-and-Localize~\cite{hu2022mix}  & 47.5 & 54.1 & 26.5 & 21.5 & 21.5 & 37.5 & 73.2 & 15.6 & 16.3 & 22.6 & 21.1 & 20.5 \\

                DSOL~\cite{hu2020dsol} & -- & -- & 30.1 & 22.3 & -- & -- & 74.3 & 15.9 & -- & -- & 22.3 & 21.1 \\

                AVGN (ours) & \textbf{50.6}	& \textbf{57.2}	& \textbf{32.5} & \textbf{24.6} & \textbf{27.3} & \textbf{42.8} & \textbf{77.5} & \textbf{18.2} & \textbf{21.9} & \textbf{28.1} & \textbf{26.2} & \textbf{23.8} \\

			\bottomrule
			\end{tabular}}
   \vspace{-0.5em}
   \caption{Quantitative results of multi-source localization on MUSIC-duet, VGGSound-Instruments, and VGGSound-Duet datasets.}
   \label{tab: exp_sota_multi}
   \vspace{-1.5em}
\end{table*}

\vspace{-0.5em}
\section{Experiments}
\vspace{-0.5em}

\subsection{Experimental setup}

\noindent\textbf{Datasets.}
MUSIC\footnote{Since many videos are no longer publicly available, the used dataset is smaller than the raw MUSIC dataset. For a fair comparison, we trained all models on the same training data.}~\cite{zhao2018the} contains 448 untrimmed YouTube music videos of solos and duets from 11 instrument categories.
358 solo videos are applied for training, and 90 solo videos are for evaluation.
124 duet videos are applied for training, and 17 duet videos are for evaluation.
Following the prior work~\cite{hu2022mix}, we use MUSIC-Solo to evaluate the performance of single-source localization, and use MUSIC-Duet for evaluating multi-source localization.
VGGSound-Instruments~\cite{hu2022mix} consists of 32k video clips of 10s lengths from 37 musical instruments categories, which is a subset of VGG-Sound~\cite{chen2020vggsound}, and each video only has a single instrument category label. 
For evaluation on multi-source localization, we follow the prior work~\cite{hu2022mix}, and randomly concatenate two frames, resulting in one input image with a shape of $448\times 224$, and summarize their waveforms for the mixture.
Beyond those musical datasets, we filter 150k video clips of 10s lengths from the original VGG-Sound~\cite{chen2020vggsound}, which is denoted as VGGSound-Single and includes 221 categories, such as nature, animals, vehicles, people, instruments, etc.
For testing, we use the full VGG-Sound Source~\cite{chen2021localizing} test set with 5158 videos for single-source localization.
For multi-source evaluation, we follow a similar setting as VGGSound-Instruments, by randomly concatenating two frames as the input image with a size of $448\times 224$, and summarizing their waveforms for the mixture.
This results in 5158 mixed videos, which is more challenging than only 446 videos in VGGSound-Instruments.
This test set is denoted as VGGSound-Duet.

\noindent\textbf{Evaluation Metrics.}
Following the prior work~\cite{hu2022mix}, we use the average precision at the pixel-wise average precision (AP), Intersection over Union (IoU), and Area Under Curve (AUC) for single-source localization.
When evaluating multi-source localization, we apply the class-aware average precision (CAP), permutation-invariant average precision (PIAP), Class-aware IoU (CIoU), and Area Under Curve (AUC) for a fair comparison with~\cite{hu2022mix}.
For the threshold of IoU and CIoU, we use IoU@0.5 and CIoU@0.3 for MUSIC-Solo and MUSIC-Duet, IoU@0.3 and CIoU@0.1 for single-source and multi-source localization on VGGSound-Instruments, IoU@0.5 and CIoU@0.3 for single-source and multi-source localization on VGGSound-Single and VGGSound-Duet.

\noindent\textbf{Implementation.}
For input images, the resolution is resized to $224 \times 224$. 
For input audio, we take the log spectrograms extracted from $3s$ of audio at a sample rate of $22050$Hz. 
We follow the prior work~\cite{mo2022EZVSL} and apply STFT to generate an input tensor of size $257 \times 300$ ($257$ frequency bands over $300$ timesteps) using 50ms windows with a hop size of 25ms. 
Following previous work~\cite{hu2019deep,qian2020multiple,chen2021localizing,mo2022EZVSL,mo2022SLAVC}, we use the lightweight ResNet18~\cite{he2016resnet} as the audio and visual encoder, and initialize the visual model using weights pre-trained on ImageNet~\cite{imagenet_cvpr09}.
$D=512$, $P=49$ for the $7\times 7$ spatial map from the visual encoder.
The depth of self-attention transformers $\phi^a(\cdot),\phi^v(\cdot)$ is 3.
The model is trained for 100 epochs using the Adam optimizer~\cite{kingma2014adam} with a learning rate of $1e-4$ and a batch size of 128.

\begin{figure*}[t]
\centering
\includegraphics[width=0.85\linewidth]{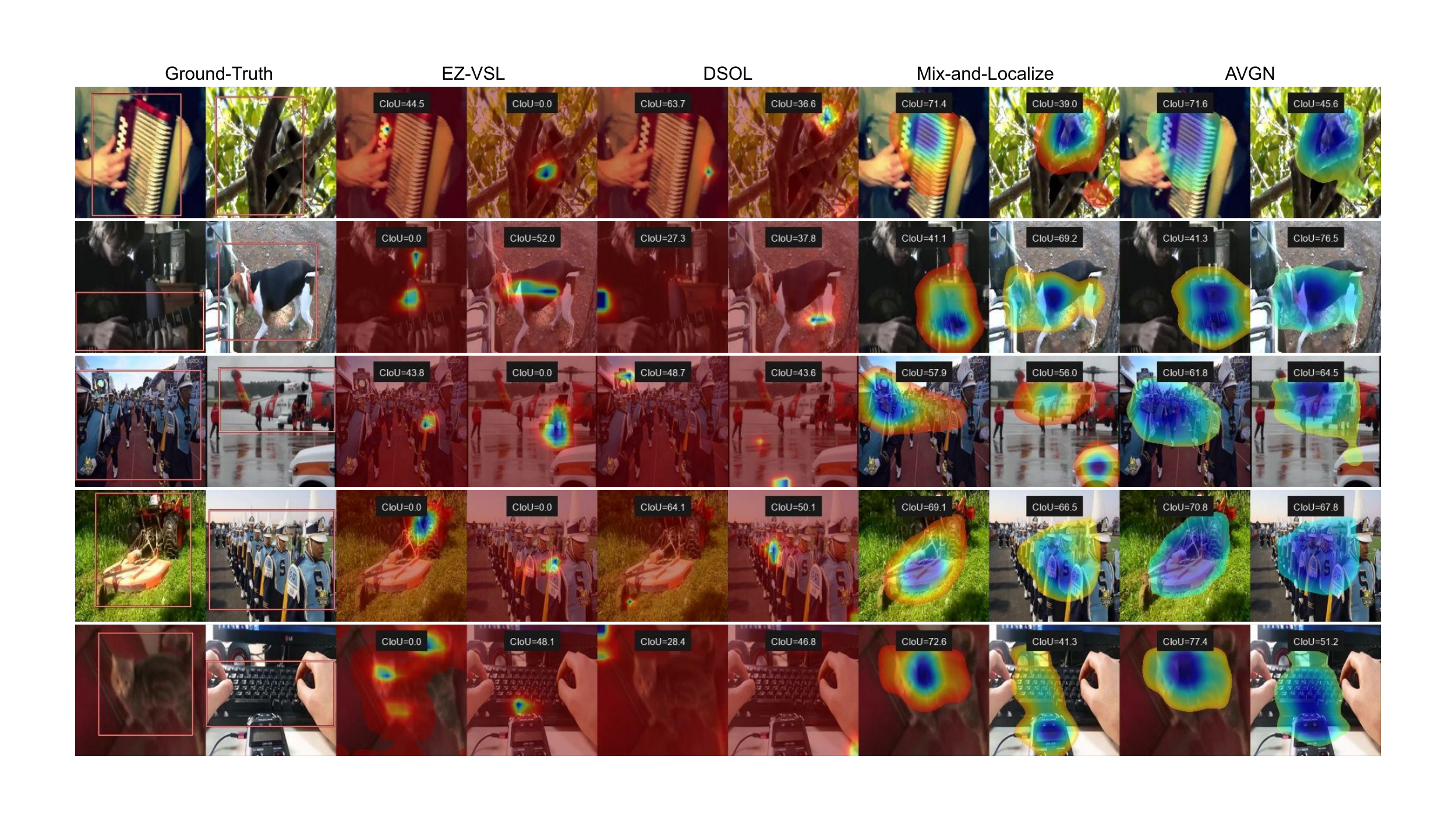}
\vspace{-1em}
\caption{Qualitative comparisons with single-source and multi-source baselines on multi-source localization. 
Note that blue color refers to high attention values and red for low attention values.
The proposed AVGN produces much more accurate and high-quality localization maps for each source. 
}
\label{fig: vis_source}
\vspace{-0.5em}
\end{figure*}

\begin{table*}[t]
	\renewcommand\tabcolsep{6.0pt}
    \renewcommand{\arraystretch}{1.1}
	\centering
	\scalebox{0.85}{
		\begin{tabular}{ccccccccc}
			\toprule
			\multirow{2}{*}{AVCT} & \multirow{2}{*}{AVG} & \multicolumn{3}{c}{MUSIC-Solo} & \multicolumn{4}{c}{MUSIC-Duet} \\
			& & AP(\%) & IoU@0.5(\%) & AUC(\%) & CAP(\%) & PIAP(\%) & CIoU@0.3(\%) & AUC(\%)  \\ 	
			\midrule
			\xmark & \xmark & 71.5 & 45.8 & 41.2 & 39.7 & 43.1 & 24.3 & 21.3 \\

			\cmark & \xmark & 75.2 & 52.3 & 45.1 & 46.9 & 51.8 & 27.6 & 22.5 \\

			\xmark & \cmark & 73.6 & 48.2 & 43.5 & 42.8 & 49.5 & 25.3 & 21.8 \\

			\cmark & \cmark & \textbf{77.2} & \textbf{58.1} & \textbf{48.5} & \textbf{50.6} & \textbf{57.2} & \textbf{32.5} & \textbf{24.6} \\
			\bottomrule
			\end{tabular}}
   \vspace{-0.5em}
   \caption{Ablation studies on Audio-Visual Class Tokens (AVCT) and Audio-Visual Grouping (AVG). }
	\label{tab: exp_ablation}
			\vspace{-1.5em}
\end{table*}

\vspace{-0.5em}
\subsection{Comparison to prior work}\label{sec:exp}
\vspace{-0.5em}

In this work, we propose a novel and effective framework for sound source localization. 
In order to validate the effectiveness of the proposed AVGN, we comprehensively compare it to previous single-source and multi-source baselines:
1) Attention 10k~\cite{Senocak2018learning}  (2018'CVPR): the first work on single-source localization with a two-stream architecture and attention mechanism;
2) OTS~\cite{Arandjelovic2018ots} (2018'ECCV): a simple baseline with audio-visual correspondence as the training objective;
3) DMC~\cite{hu2019deep} (2019'CVPR): a deep multi-modal clustering network with audio-visual co-occurrences to learn convolutional maps for each modality in different embedding spaces;
4) CoarsetoFine~\cite{qian2020multiple} (2020'ECCV): a two-stage baseline with coarse-to-fine alignment for cross-modal features;
5) DSOL~\cite{hu2020dsol} (2020'NeurIPS): a two-stage training framework with classes as weak supervision for category-aware sound source localization;
6) LVS~\cite{chen2021localizing} (2021'CVPR): a contrastive network to learn audio-visual correspondence maps with hard negative mining;
7) EZ-VSL~\cite{mo2022EZVSL} (2022'ECCV): a recent strong baseline with multiple-instance contrastive learning for single-source localization;
8) Mix-and-Localize~\cite{hu2022mix} (2022'CVPR): a strong multi-source baseline using a contrastive random walk algorithm in the graph composed of images and separated sounds as nodes.

For single-source localization, we report the quantitative comparison results in Table~\ref{tab: exp_sota_single}.
As can be seen, we achieve the best performance in terms of all metrics for three benchmarks, compared to previous self-supervised and weakly-supervised baselines.
In particular, the proposed AVGN significantly outperforms DSOL~\cite{hu2020dsol}, the current state-of-the-art weakly-supervised baseline, by 6.7 IoU@0.5 \& 4.8 AUC, 5.1 IoU@0.5 \& 3.8 AUC, and 5.1 IoU@0.5 \& 5.1 AUC on three datasets.
Moreover, we achieve superior performance gains compared to EZ-VSL~\cite{mo2022EZVSL}, the current state-of-the-art self-supervised baseline, which implies the importance of extracting category-aware semantics from audio-visual inputs as the guidance for learning audio-visual alignment discriminatively.
Meanwhile, our AVGN outperforms than Mix-and-Localize~\cite{hu2022mix} by a large margin, where we achieve the performance gains of 8.6 AP on MUSIC-Solo, 5.6 AP on VGGSound-Instruments, and 2.7 AP on VGGSound-Single.
These significant improvements demonstrate the superiority of our method in single-source localization.

In addition, significant gains in multi-source sound localization can be observed in Table~\ref{tab: exp_sota_multi}.
Compared to Mix-and-Localize~\cite{hu2022mix}, the current state-of-the-art multi-source localization baseline, we achieve the results gains of 5.8 CAP, 5.3 PIAP, 4.3 CIoU@0.1, and 2.6 AUC on VGGSound-Instruments.
Furthermore, when evaluated on the challenging VGGSound-Duet benchmark, the proposed approach still outperforms Mix-and-Localize~\cite{hu2022mix} by 5.6 CAP, 5.5 PIAP, 5.1 CIoU@0.3, and 3.3 AUC.
We also achieve highly better results against DSOL~\cite{hu2020dsol}, the weakly-supervised baseline with two training stages.
These results validate the effectiveness of our approach in learning disentangled individual source semantics from mixtures and images for multi-source localization.

In order to qualitatively evaluate the localization maps, we compare the proposed AVGN with EZ-VSL~\cite{mo2022EZVSL}, Mix-and-Localize~\cite{hu2022mix}, and DSOL~\cite{hu2020dsol} on both single-source and multi-source localization in Figure~\ref{fig: vis_source}.
From comparisons, three main observations can be derived:
1) Without explicit separation objectives, EZ-VSL~\cite{mo2022EZVSL}, the strong single-source baseline, performs worse on the multi-source localization.
2) the quality of localization maps generated by our method is much better than the self-supervised multi-source baseline, Mix-and-Localize~\cite{hu2022mix}.
3) the proposed AVGN achieves competitive even better results on predicted maps against the weakly-supervised multi-source baseline~\cite{hu2020dsol} by using category labels during training.
These visualizations further showcase the superiority of our simple AVGN in learning category-aware audio-visual representations to guide localization for each source.

\vspace{-0.5em}
\subsection{Experimental analysis}
\vspace{-0.5em}

In this section, we performed ablation studies to demonstrate the benefit of introducing the Audio-Visual Class Tokens and Audio-Visual Grouping module. 
We also conducted extensive experiments to explore a flexible number of sound source localization, and learned disentangled category-aware  audio-visual representations.

\begin{figure*}[t]
\centering
\includegraphics[width=0.8\linewidth]{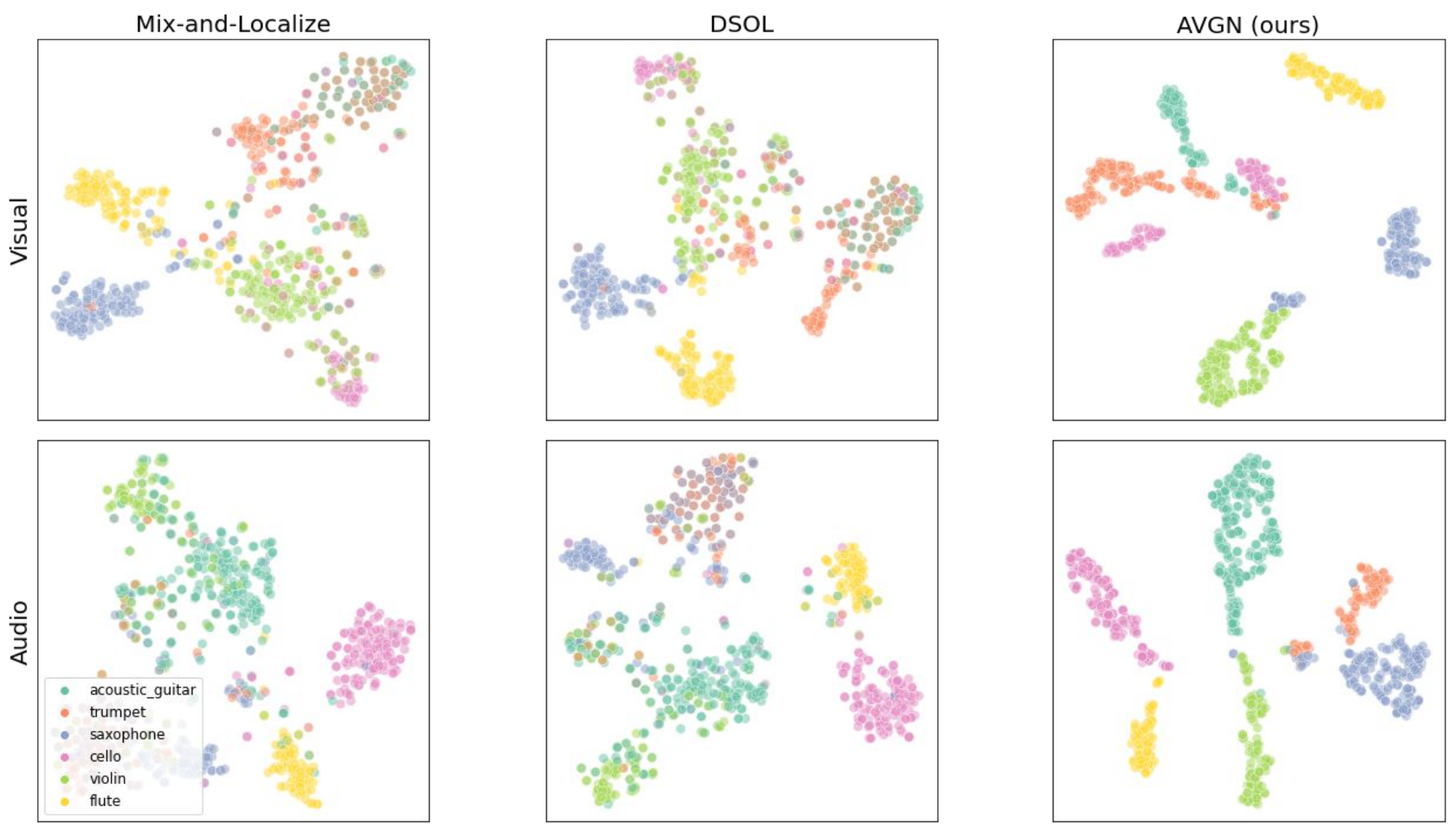}
\vspace{-1em}
\caption{Qualitative comparisons of representations learned by Mix-and-Localize, DSOL, and the proposed AVGN. 
Note that each spot denotes the feature of one source sound, and each color refers to one source category, such as ``trumpet" in orange and ``cello" in pink.}
\label{fig: exp_vis_feat}
\vspace{-1.5em}
\end{figure*}

\noindent\textbf{Audio-Visual Class Tokens \& Audio-Visual Grouping.}
In order to validate the effectiveness of the introduced audio-visual class tokens (AVCT) and audio-visual grouping (AVG), we ablate the necessity of each module and report the quantitative results in Table~\ref{tab: exp_ablation}.
We can observe that adding bearable AVCT to the vanilla baseline highly increases the results of single-source localization (by 3.7 AP, 6.5 IoU@0.5, and 3.9 AUC) and multi-source localization (by 7.2 CAP, 8.7 PIAP, and 3.3 CIoU@0.3), which demonstrates the benefit of category tokens in extracting disentangled high-level semantics for source localization. 
Meanwhile, introducing only AVG in the baseline also increases the source localization performance in terms of all metrics.
More importantly, incorporating AVCT and AVG together into the baseline significantly raises the performance by 5.7 AP, 12.3 IoU@0.5 and 7.3 AUC on single-source, and by 10.9 CAP, 14.1 PIAP, 8.2 CIoU@0.3 and 3.3 AUC on multi-source localization. 
These improving results validate the importance of audio-visual class tokens and audio-visual grouping in extracting category-aware semantics from the mixture and image for sound localization.

\noindent\textbf{Generalizing to Flexible Number of Sources.}
In order to show the generalizability of the proposed AVGN to a flexible number of sources, we directly transfer the model without additional training to test a mixture of 3 sources.
We still achieve competitive results of 18.5 CAP, 23.7 PIAP, 22.7 CIoU@0.3, and 21.8 AUC on the challenging VGGSound-Duet dataset.
These results indicate that our AVGN can support localizing a flexible number of sources from the mixture, which is different from Mix-and-Localize~\cite{hu2022mix} with a fixed number of sources as the number of nodes defined in the trained contrastive random walker.

\noindent\textbf{Learned Category-aware Audio-Visual Representations.}
Learning disentangled audio-visual representations with category-aware semantics is critical for us to localize the sound source from a mixture.
To better evaluate the quality of learned category-aware features, we visualize the learned visual and audio representations of 6 categories in MUSIC-Duet by t-SNE~\cite{laurens2008visualizing}, as shown in Figure~\ref{fig: exp_vis_feat}.
Note that each color refers to one category of the source sound, such as ``trumpet'' in orange and ``cello'' in pink.
As can be observed in the last column, audio-visual representations extracted by the proposed AVGN are both intra-category compact and inter-category separable. 
In contrast to our disentangled embeddings in the audio-visual semantic space, there still exists mixtures of multiple audio-visual categories among features learned by Mix-and-Localize~\cite{hu2022mix}.
With the benefit of the weakly-supervised classes, DSOL~\cite{hu2020dsol} can extract clustered audio-visual features on some classes, such as ``cello'' in pink. 
However, most categories are mixed together as they do not incorporate the explicit audio-visual grouping mechanism in our AVGN.
These meaningful visualization results further showcase the success of our AVGN in extracting compact audio-visual representations with class-aware semantics for sound source localization from the mixture.

\vspace{-0.5em}

\subsection{Limitation}

Although the proposed AVGN achieves superior results on both single-source and multi-source localization, the performance gains of our approach on the MUSIC-Duet benchmark with a small number of categories are not significant.
One possible reason is that our model easily overfits across the training phase, and the solution is to incorporate dropout and momentum encoders together for multi-source localization.
Meanwhile, we notice that if we transfer our model to open-set source localization without additional training, it would be hard to localize unseen categories as we need to pre-define a set of categories during training and do not learn unseen category tokens to guide multi-source localization.
The future work could be to add enough learnable class tokens or apply continual learning to new classes.

\vspace{-0.5em}
\section{Conclusion}


In this work, we present AVGN, a novel audio-visual grouping network, that can directly learn category-wise semantic features for each source from audio and visual inputs for localizing multiple sources in videos.
We introduce learnable audio-visual category tokens to aggregate class-aware source features.
Then, we leverage the aggregated semantic features for each source to guide localizing the corresponding regions.
Compared to existing multi-source methods, our new framework can handle a flexible number of sources and learns compact audio-visual semantic representations.
Empirical experiments on MUSIC, VGGSound-Instruments, and VGG-Sound Sources benchmarks demonstrate the state-of-the-art performance of our AVGN on both single-source and multi-source localization.

\noindent\textbf{Broader Impact.}
The proposed method localizes sound sources learning from user-uploaded web videos, which might cause the model to learn internal biases in the data.
For example, the model could fail to localize certain rare but crucial sound sources.
These issues should be carefully addressed for the deployment of real scenarios.

\newpage

{\small
\bibliographystyle{ieee_fullname}
\bibliography{reference}
}

\newpage

\appendix
\section*{Appendix}

In this appendix, we provide the significant differences between our AVGN and the recent work, GroupViT~\cite{xu2022groupvit}, more experiments on the depth of transformer layers and grouping strategies. 
In addition, we validate the effectiveness of learnable audio-visual class tokens in learning disentangled audio-visual representations and report qualitative visualization results of localization maps.

\begin{table*}[t]
	\renewcommand\tabcolsep{6.0pt}
    \renewcommand{\arraystretch}{1.1}
	\centering
	\scalebox{0.85}{
		\begin{tabular}{ccccccccc}
			\toprule
			\multirow{2}{*}{Depth} & \multirow{2}{*}{AVG} & \multicolumn{3}{c}{MUSIC-Solo} & \multicolumn{4}{c}{MUSIC-Duet} \\
			& & AP(\%) & IoU@0.5(\%) & AUC(\%) & CAP(\%) & PIAP(\%) & CIoU@0.3(\%) & AUC(\%)  \\ 	
			\midrule
			1 & Softmax & 75.8 & 53.6 & 45.7 & 47.6 & 53.1 & 28.5 & 23.2 \\
                3 & Softmax & \textbf{77.2} & \textbf{58.1} & \textbf{48.5} & \textbf{50.6} & \textbf{57.2} & \textbf{32.5} & \textbf{24.6} \\
                
                6 & Softmax & 76.7 & 57.5 & 47.9 & 50.1 & 56.9 & 32.1 & 24.3 \\

                12 & Softmax & 76.3 & 57.3 & 47.6 & 49.8 & 56.7 & 31.7 & 24.1 \\

                3 & Hard-Softmax & 73.2 & 47.6 & 43.1 & 42.5 & 49.2 & 24.8 & 21.5 \\
			\bottomrule
			\end{tabular}}
   \vspace{-0.5em}
   \caption{Exploration studies on the depth of self-attention transformer layers and grouping strategies in Audio-Visual Grouping (AVG) module. }
	\label{tab: ab_depth}
			\vspace{-1.5em}
\end{table*}

\begin{figure*}[t]
\centering
\includegraphics[width=0.9\linewidth]{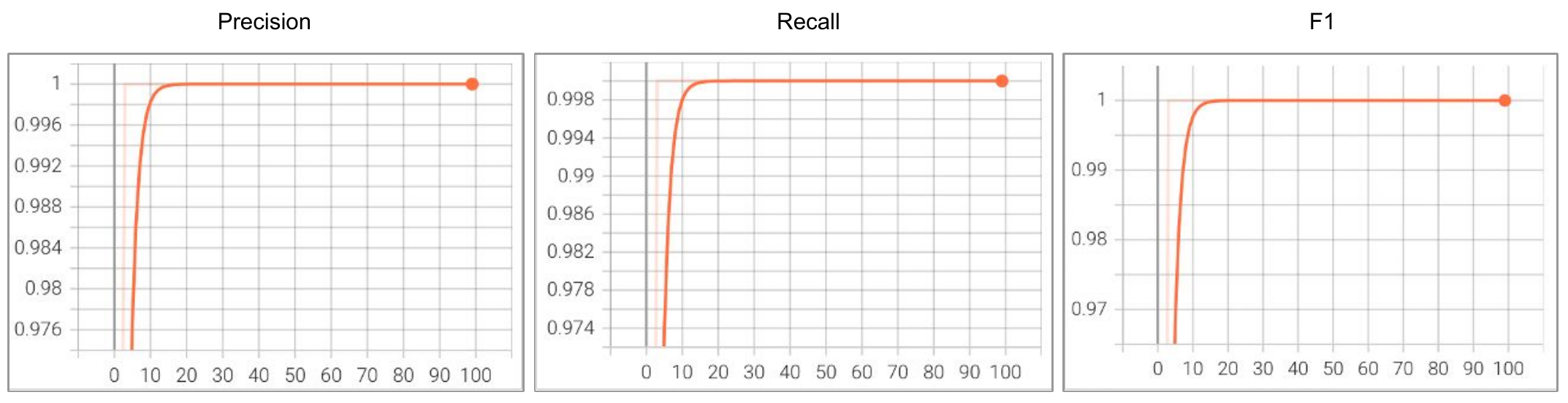}
\caption{Quantitative results (Precision, Recall, and F1 score) of learned audio-visual class tokens.}
\label{fig: vis_curve_cls}
\end{figure*}

\begin{figure*}[!hbt]
\centering
\includegraphics[width=0.98\linewidth]{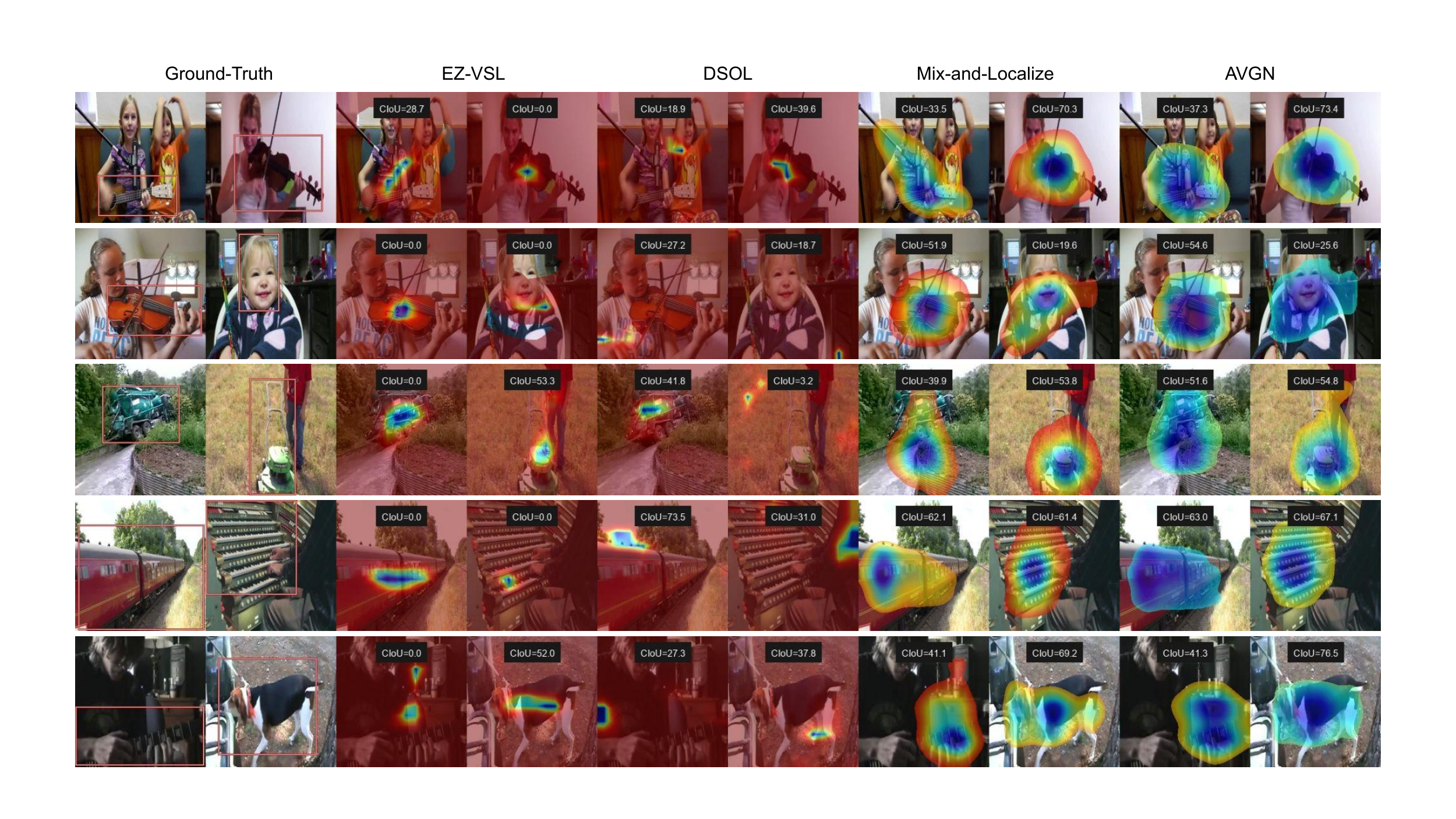}
\caption{Qualitative comparisons with single-source and multi-source baselines on multi-source localization.
Note that blue refers to high attention values and red for low attention values.
The proposed AVGN produces more accurate and high-quality localization maps for each source.
}
\label{fig: exp_vis_suppl}
\end{figure*}

\section{Significant Difference from GroupViT and AVGN}

When compared to GroupViT~\cite{xu2022groupvit} on image segmentation, there are three significant distinct characteristics of our AVGN for addressing sound localization from mixtures, which are highlighted as follows:

1) \textbf{Constraint on Audio-Visual Category Tokens.}
The major difference is that we have learned disentangled audio-visual class tokens for each sound source, \textit{e.g.}, 37 audio-visual category tokens for 37 categories in the VGGSound-Instruments benchmark.
During training, each audio-visual class token does not learn semantic overlapping information among each other, where we apply the cross-entropy loss $\sum_{i=1}^C\mbox{CE}(\mathbf{h}_i, \mathbf{e}_i)$ on each category probability $\mathbf{e}_i$ with the disentangled constraint $\mathbf{h}_i$.
However, the number of group tokens used in GroupViT is a hyper-parameter, and they must tune it carefully across each grouping stage.

2) \textbf{Audio-Visual Grouping.}
We propose the audio-visual grouping module for extracting individual semantics with category-aware information from the mixture spectrogram and image. 
However, GroupViT used the grouping mechanism on only visual patches without explicit category-aware tokens involved. 
Therefore, GroupViT can not be directly applied to a sound spectrogram for solving sound localization problems from mixtures. 
Moreover, they utilized multiple grouping stages during training and the number of grouping stages is a hyper-parameter. 
In our module, only one audio-visual grouping stage with disentangled audio-visual category tokens is enough to learn disentangled audio-visual representations in the multi-modal semantic space.

3) \textbf{Audio-Visual Class as Weak Supervision.}
We leverage the audio-visual category as the weak supervision to address the sound localization problem from the mixtures, while GroupViT used a trivial contrastive loss to match the global visual representations with text embeddings. 
In this case, GroupViT required a large batch size for self-supervised training on large-scale visual-language pairs. 
In contrast, we do not need unsupervised learning on the large-scale simulated mixture data with extensive training costs.

\section{Depth of Transformer Layers and Grouping Strategies}

The depth of transformer layers and grouping strategies used in the proposed AVG affect the extracted and grouped representations for source localization from the mixtures.
To explore such effects more comprehensively, we varied the depth of transformer layers from $\{1, 3, 6, 12\}$ and ablated the grouping strategy using Softmax and Hard-Softmax.
During training, the Gumbel-Softmax~\cite{eric2017categorical,chris2017the} was applied as the alternative to Hard-Softmax to make it differentiable.

We report the comparison results of source localization performance in Table~\ref{tab: ab_depth}.
When the depth of transformer layers is 3 and using Softmax in AVG, we achieve the best localization performance in terms of all metrics. 
With the increase of the depth from 1 to 3, the proposed AVGN consistently raises results as better disentangled audio-visual representations are extracted from encoder embeddings of the raw mixture and image. 
However, increasing the depth from 3 to 12 will not continually improve the result since 3 transformer layers might be enough to extract the learned category-aware embeddings for audio-visual grouping with only one grouping stage.
Furthermore, replacing Softmax with Hard-Softmax significantly deteriorates the localization performance, which shows the importance of the proposed AVG in extracting disentangled audio-visual representations with category-aware semantics from the audio mixture and image for localizing each sounding source.

\section{Quantitative Validation on Audio-Visual Category Tokens}

Learnable Audio-Visual Category Tokens are essential to aggregate audio-visual representations with category-aware semantics from the sound mixture.
To quantitatively validate the rationality of learned audio-visual category token embeddings, we compute the Precision, Recall, and F1 scores of audio-visual classification using these embeddings across training iterations. The quantitative results are reported in Figure~\ref{fig: vis_curve_cls}.
As can be seen, all metrics rise to 1 at epoch 20, which indicates that each learned audio-visual category token has disentangled information with category-aware semantics. 
These quantitative results further demonstrate the effectiveness of audio-visual category tokens in the audio-visual grouping for extracting disentangled audio-visual representations from images and audio mixtures for localizing each source more accurately.

\section{Qualitative Visualization on Source Localization}

To qualitatively demonstrate the effectiveness of our method, we report more visualization results in Figure~\ref{fig: exp_vis_suppl}.
We can observe that the proposed AVGN achieves decent localization performance in terms of more accurate and high-quality localization maps for each sound source.

\end{document}